\documentclass[10pt,twocolumn,letterpaper]{article}

\usepackage{iccv}
\usepackage{times}
\usepackage{epsfig}
\usepackage{graphicx}
\usepackage{amsmath}
\usepackage{amssymb}
\usepackage{caption}
\usepackage{multirow}
\usepackage{enumitem}
\usepackage[pagebackref=true,breaklinks=true,letterpaper=true,colorlinks,bookmarks=false]{hyperref}

\iccvfinalcopy % *** Uncomment this line for the final submission

\def\iccvPaperID{5892} % *** Enter the ICCV Paper ID here

\ificcvfinal\fi
\begin{document}

%%%%%%%%% TITLE
\title{Graph-based Spatial-temporal Feature Learning for Neuromorphic Vision Sensing}

\author{Yin Bi, Aaron Chadha, Alhabib Abbas, Eirina Bourtsoulatze and Yiannis Andreopoulos\\
Department of Electronic \& Electrical Engineering\\
University College London, London, U.K.\\
{\tt\small \{yin.bi.16, aaron.chadha.14, alhabib.abbas.13, e.bourtsoulatze, i.andreopoulos\}@ucl.ac.uk}
% For a paper whose authors are all at the same institution,
% omit the following lines up until the closing ``}''.
% Additional authors and addresses can be added with ``\and'',
% just like the second author.
% To save space, use either the email address or home page, not both
%\and
%Second Author\\
%Institution2\\
%First line of institution2 address\\
%{\tt\small secondauthor@i2.org}
%\thanks{This work was funded by EPSRC, grants EP/R025290/1 and EP/P02243X/1, and European Union�s Horizon 2020 research and innovation  programme (Marie Sklodowska-Curie fellowship, grant agreement No. 750254).}
}

\maketitle
%\thispagestyle{empty}

%%%%%%%%% ABSTRACT
\begin{abstract}
Neuromorphic vision sensing (NVS)\ devices represent visual information as sequences of asynchronous discrete events (a.k.a., ``spikes'') in response to  changes in scene reflectance.
Unlike conventional active pixel sensing (APS), NVS allows for significantly higher event sampling rates at substantially increased energy efficiency and robustness to illumination changes. However, feature representation for NVS is far behind its APS-based counterparts, resulting in lower performance in high-level computer vision tasks. To fully utilize its sparse and asynchronous nature, we propose a compact graph representation for NVS, which allows for end-to-end learning with graph convolution neural networks. We couple this with a  novel end-to-end feature learning framework that accommodates both appearance-based and motion-based tasks. The core of our framework comprises a spatial feature learning module, which utilizes residual-graph convolutional neural networks (RG-CNN), for end-to-end learning of appearance-based features directly from  graphs. We extend this with our proposed  Graph2Grid  block and temporal feature learning module for efficiently modelling temporal dependencies over multiple graphs and a long temporal extent. We show how our framework can be configured for  object classification, action recognition and action similarity labeling. Importantly, our approach preserves the spatial and temporal coherence of spike events, while requiring less computation and memory. The experimental validation shows that our proposed framework outperforms all recent methods on standard datasets.  Finally, to address the absence of large real-world NVS datasets for complex recognition tasks,  we introduce, evaluate  and make available the American Sign Language letters (ASL-DVS), as well as human action dataset (UCF101-DVS, HMDB51-DVS and ASLAN-DVS).

\end{abstract}

%%%%%%%%% BODY TEXT
%%%%%%%%% BODY TEXT
\section{Introduction}

\begin{figure}
\centering
\vspace{-0.1in}
\includegraphics[width=3.2in]{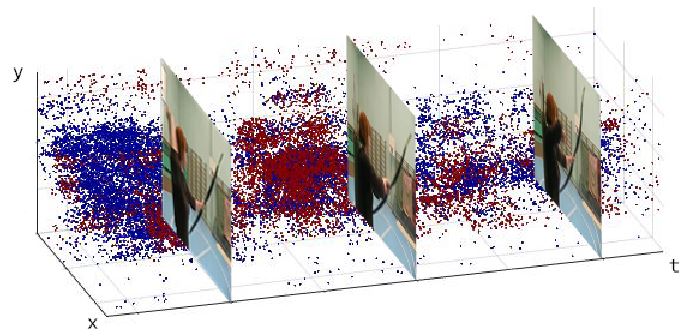}
\vspace{-0.1in}
\caption{Examples of archery action captured by APS and NVS sensors. APS sensors capture images at fixed frame rates, while NVS sensors output a stream of events. (Red:ON, Blue:OFF)}
\vspace{-0.1in}
\label{f:example}
\end{figure}

With the prevalence and advances of CMOS active pixel sensing (APS) and deep learning, researchers have achieved good performance in APS-based computer vision tasks, such as object detection \cite{girshick2015fast, ren2015faster}, object recognition  \cite{krizhevsky2012imagenet, deng2009imagenet} and action recognition \cite{tran2015learning, carreira2017quo} . However, APS cameras suffer from limited frame rate, high redundancy between frames, blurriness due to slow shutter adjustment under varying illumination, and high power requirements \cite{tobi2016neu} which limit the effectiveness of APS-based frameworks. To solve these problems, researchers have devised neuromorphic vision sensing (NVS) sensors such as  the iniLabs DAVIS cameras \cite{delbruck2010activity} and the Pixium Vision ATIS cameras \cite{posch2010qvga}, which are  inspired by the photoreceptor-bipolar-ganglion cell information flow in mammalian vision. NVS sensors generate output (i.e., spikes)  asynchronously only when  the transient change of illumination intensity in a scene exceeds a certain threshold, instead of recording entire frames at fixed frame rates, independent of any activity in the scene (as per APS sensors). The output of the NVS sensor is represented asynchronously as a collection of tuple sequences, referred to as an Address Event Representation (AER)\cite{berner20075}, which contains the spatio-temporal coordinates of the  reflectance events along with the event polarity (i.e., ON or OFF).  The event polarity indicates an increase (ON) or decrease (OFF) in illumination intensity.  As an illustration, Fig. \ref{f:example} shows a neuromorphic event stream, overlaid with  the corresponding  RGB frames recorded at the video framerate; events  are plotted according to their spatio-temporal coordinates and color coded as blue (OFF) and red (ON). Notably, there are many more intermediate events between the RGB frames, which indicates the substantially higher framerate achievable with an NVS sensor and asynchronous  outputs. Furthermore,  the asynchronicity  removes the data redundancy from the scene, which reduces to the power requirement to 10mW, compared to several hundreds of mW for APS sensors. Remarkably, NVS sensors achieve this with microsecond-level latency and robustness to uncontrolled lighting conditions as no synchronous global shutter is used.

Beyond event sparsity and asynchronicity, neuromorphic event streams are naturally encoding spatio-temporal motion information \cite{tobi2016neu}; as such, they  are extremely adaptable to tasks related to moving objects such as action analysis/recognition, object tracking or high-speed moving scenes. We, therefore, look to perform feature learning directly on the raw neuromorphic events. %This is in contrast to recent work \cite{feichtenhofer2016convolutional}, which relies on extracting spatio-temporal features from RGB frames and, thus, inherits the limitations associated with APS cameras.     
Unfortunately, effective methods for representation learning on neuromorphic events to solve complex computer vision tasks are currently limited and outperformed by their  APS-based counterparts.  This is partly due to a limited research  in the NVS domain, as well as a lack of NVS data with reliable annotations to train and test on \cite{tobi2016neu,tan2015benchmarking}. Yet, more so, the sheer  abundance of asynchronous and sparse events means that feature learning directly on events can be particularly cumbersome and unwieldy. Thus far, most approaches have attempted to solve this issue by either artificially grouping events into frame forms  \cite{zhu2018ev, cannici2018event} or deriving complex feature descriptors \cite{sironi2018hats, lagorce2017hots}, which do not always provide for good representations for complex tasks like object classification. Moreover, such approaches dilute the advantages of the  asynchronicity of NVS streams by limiting the frame-rate, and may be sensitive to the noise and change of camera motion or viewpoint orientation. Finally, these methods fail to model long temporal event dependencies explicitly, thus rendering them less viable for motion-based tasks. 

More recent methods on feature representation have employed end-to-end feature learning, where a convolutional neural network (CNN) \cite{ghosh2019spatiotemporal, amir2017low}  or spiking neural network (SNN) \cite{diehl2015unsupervised, lee2016training} is trained to learn directly from raw observations. While these methods show great promise, CNN-based learning methods require event grouping into frames and, therefore, suffer from the same  drawbacks as above. On the other hand, SNN-based methods are complex to train, which results in lower performance compared to gradient-based alternatives. Instead of using CNNs or SNNs, we propose to leverage  on graph-based learning, by training an end-to-end feature learning framework directly on neuromorphic events.  By representing events as  graphs, we are able to maintain event asynchronocity and sparsity, while performing training with traditional gradient-based backpropagation. To the best of our knowledge, this is the first attempt to represent neuromorphic spike events as graphs, which allows to use graph convolutional neural networks for end-to-end feature learning directly on neuromorphic events. Building partly on our previous work \cite{bi2019graph}, our proposed graph based framework is able to accommodate both appearance  and motion-based tasks; in this paper, we focus on  object classification, action recognition and action similarity labelling as representative tasks. For object classification, we design a spatial feature learning module, comprising graph convolutional layers and graph pooling layers for processing a single input event graph. For action recognition and action similarity labeling, we extend this module  with temporal feature learning, in order to learn a spatio-temporal representation over the entire input. Specifically, we introduce a Graph2Grid block for aggregating a sequence of graphs over a long temporal extent. Each event graph in the sequence is first processed by a spatial feature learning module; the mapped graphs are then converted to grid representation by the Graph2Grid block and the resulting frames are stacked, for processing with any conventional 2D or 3D CNNs. This is inspired by recent work in APS-based action recognition \cite{feichtenhofer2016convolutional} that processes multiple RGB frames with 2D CNNs and aggregates the learned representations with a 3D convolution fusion and pooling.

In order to address the lack of NVS data for evaluation, we introduce the largest sourced NVS dataset for object classification, which we refer to as ASL-DVS. The task is to classify hand recordings  as one of 24 letters from the American Sign Language (ASL).  For action recognition and action similarity labeling, we leverage existing APS-based datasets such as UCF101 \cite{soomro2012ucf101}, HMDB51 \cite{kuehne2011hmdb} and ASLAN \cite{kliper2011action}, and convert these to the NVS domain  by  recording a playback of each dataset captured from a display with a DAVIS240c NVS camera. The generated NVS datasets, UCF101-DVS, HMDB51-DVS and ASLAN-DVS, include more content than any previous NVS dataset in these action-based tasks.

We evaluate our framework on object classification, action recognition and action similarity labelling, and show that our framework achieves state-of-the-art results on both tasks compared to recent work on conventional frame-based approaches. We summarize our contributions as follows:

\begin{enumerate} [leftmargin=*]
\item We propose a novel graph based representation for neuromorphic events, allowing for fast end-to-end graph based training and inference;
\item We design a new graph-based spatial feature learning module and evaluate its performance on  object classification;  
\item We extend  our spatial feature learning module with our Graph2Grid block and temporal feature learning module for efficiently modelling coarse temporal dependencies over multiple graphs. We evaluate performance of the learning framework on  action recognition and action similarity labeling. 
\item We introduce new datasets for NVS-based object classification (ASL-DVS), action recognition (UCF101-DVS and HMDB51-DVS) and action similarity labeling (ASLAN-DVS) to address the lack of NVS data for training and inference, and make these available to the research community. 
\end{enumerate}

In Section \ref{sec:related_work} we review related work. Section \ref{sec:methods} details our method for graph-based spatio-temporal feature learning network.  Three downstream applications including object classification, human action recognition and action similarity labeling are presented in Section \ref{sec:applications}. Section \ref{sec:conclusion} concludes the paper.

\section{Related Work}\label{sec:related_work}

In the field of neuromorphic vision,  recent literature focuses on two types of feature representation: handcrafted feature extraction and end-to-end trainable feature learning. Handcrafted feature descriptors are widely used by neuromorphic vision community. Some of the most common  are corner detectors and line/edge extraction \cite{mueggler2017fast, mueggler2017event}. While these efforts were promising early attempts for NVS-based object classification, their performance does not scale well when considering complex datasets. Inspired by their frame-based counterparts, optical flow methods have been proposed as feature descriptors for NVS \cite{clady2017motion, benosman2014event}. For a high-accuracy optical flow, these methods have very high computational requirements, which  diminishes their usability in real-time applications. In addition, due to the inherent discontinuity and irregular sampling of NVS data, deriving compact optical flow representations with enough descriptive power for accurate classification and tracking still remains a challenge \cite{clady2017motion}.  %Orchard  \textit{et al.} introduced HFirst descriptors that used spike timing to encode the strength of events and implemented a max operation to output a number representing the strength input \cite{orchard2015hfirst}.
Lagorce \textit{et al.} proposed event based spatio-temporal features called time-surfaces \cite{lagorce2016hots}. This is a time oriented approach to extract spatio-temporal features that are dependent on the direction and speed of motion of the objects. Inspired by time-surfaces, Sironi \textit{et al.} proposed a higher-order representation for local memory time surfaces that emphasizes the importance of using the information carried by past events to obtain a robust representation \cite{sironi2018hats}.% Recently Ramesh  \textit{et al.} introduced a generic visual descriptor termed as DART that encodes the structural context using log-polar grids for events \cite{ramesh2019dart}. 
These descriptors are very sensitive to noise and strongly depend on the type of object motion in scene. Moreover, they fail to take temporal information into account and maintain a representation of dynamics over a long time. Thus, they can only be used for static object recognition, and not for long temporal applications such as action recognition evaluated in this work. 

End-to-end feature learning for NVS-based tasks consists of two types of approaches: frame-based and event-based. The main idea of frame-based methods is to convert the neuromorphic events into synchronous frames of spike events, on which conventional computer vision techniques can be applied for the feature learning.  Zhu  \textit{et al.} \cite{zhu2018ev} introduced a four-channel image form with the same resolution as the neuromorphic vision sensor. % the first two channels encode the number of positive and negative events that have occurred at each pixel, while last two channels as the timestamp of the most recent positive and negative event. 
Inspired by the functioning of spiking neural networks (SNNs) to maintain memory of past events, leaky frame integration has been used in recent work \cite{cannici2018event, cannici2018attention}, where the corresponding position of the frame is incremented by a fixed amount when a event occurs at the same event address. %Peng  \textit{et al.} \cite{peng2017bag} proposed bag-of-events (BOE) feature descriptors, which is a statistical learning method that firstly divides the event streams into multiple segments and then relies on joint probability distribution of the consecutive events to represent feature maps. 
Amir \textit{et al.} use a cascade of temporal filters to process the events, which is regarded as stacking frames, and then feed these frames into  a CNN \cite{amir2017low}. Similary, Ghosh  \textit{et al.} partitioned events into a three-dimensional grid of voxels where spatio-temporal filters are used to learn the features, and learnt features are fed as input to CNNs for action recognition \cite{ghosh2019spatiotemporal}. Chadha  \textit{et al.} \cite{chadha2019neuromorphic} generated frames by summing the polarity of events in each address as pixel, then fed them into a multi-modal teacher-student framework for action recognition. % where it employs a pre-trained optical flow stream as a teacher network to transfer knowledge to the NVS student network . 
While useful for early-stage attempts, these frame-based methods are not well-suited for the neuromorphic event's sparse and asynchronous nature since the frame sizes that need to be processed are substantially larger than those of the original NVS streams. The advantages of event-based sensors are diluted if their event streams are cast back into synchronous frames for the benefit of conventional processors downstream, thus not providing efficient and power-saving learning systems. 

The second type of end-to-end feature learning methods are event-based methods. The most commonly used architecture relies on spiking neural networks (SNNs)  \cite{akopyan2015truenorth, diehl2015unsupervised, lee2016training} for inference. While SNNs are theoretically capable of learning complex representations, they  still fail the performance of gradient-based methods due to the lack
of suitable training algorithms. Essentially, since the activation functions of spiking neurons are not differentiable, SNNs are not able to leverage on popular training methods such as backpropagation. To address this, researchers currently follow a hybrid approach \cite{diehl2015fast, stromatias2015scalable}: a neural network is trained off-line using continuous/rate-based neuronal models with state-of-the-art supervised training algorithms; then, the trained architecture is mapped to an SNN. However, until now, despite their substantial implementation advantages at inference, 
the obtained solutions are complex to train and typically achieve lower performance than gradient-based CNNs. Thus, other directions for event-based feature learning for neuromorphic vision sensing have been also explored. Wang \textit{et al.} interpreted an event sequence as a 3D point cloud in space and time\cite{wang2019space}, which is hierarchically fed into PointNet\cite{qi2017pointnet} to capture the spatio-temproal structure of motion. While providing useful insights, all these methods were tested on simple datasets (e.g., the DVS128 Gesture dataset \cite{amir2017low} of gestures and postures) with a small number of classes and clean background. It is, therefore, unlikely that these methods can obtain such high accuracy for real-world scenarios, as they cannot capture long-term temporal dependencies. When applied to complex datasets (e.g., UCF101\_DVS) for human action recognition, the performance of these methods degrades significantly.  %: e.g., DVS128 Gesture Dataset used in \cite{amir2017low, wang2019space} comprises a set of 11 hand and arm gestures and the posture dataset in \cite{zhao2014feedforward, peng2016bag} includes only three human actions, i.e., bend, sitsand, and walk.  Moreover, all current datasets all were acquired from a relatively noise-free experimental environment that cannot represent complex real-life scenarios, which led to many algorithms achieving very high accuracy. It is therefore unlikely that these methods cannot obtain such high accuracy for real-world scenarios, as they cannot capture long-term temporal dependencies. When these work on complex dataset (such as UCF101-DVS) for human action recognition, their performance degrades significantly as discussed in application experiments.

\section{Methodology}\label{sec:methods}

\begin{figure*}
  \vspace{-0.25in}
  \centering
  \includegraphics[width=7.0in]{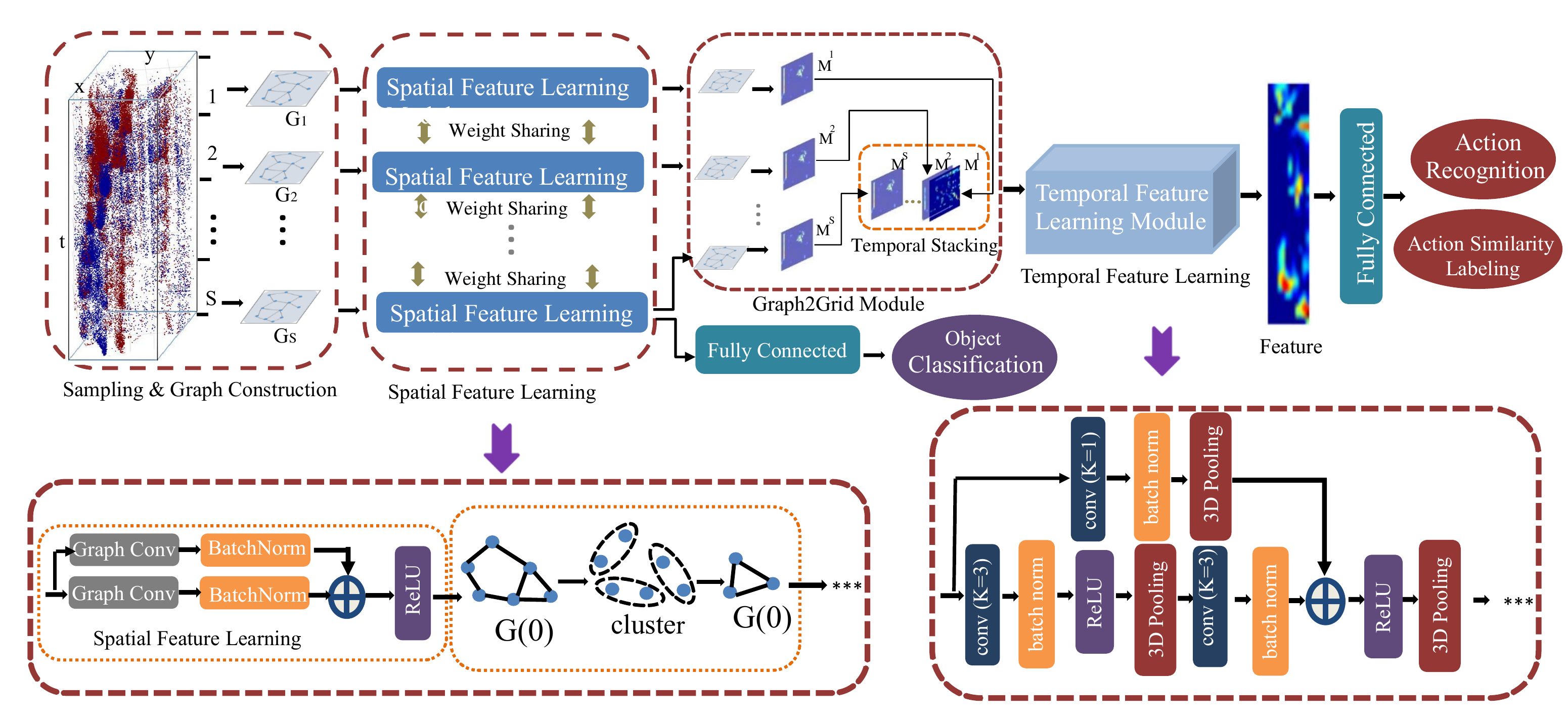}
  \vspace{-0.05in}
  \caption{Framework of graph-based spatial-temporal feature learning for neuromorphic vision sensing. Our framework is able to accommodate both object classification and action recognition/similarity labeling tasks.We first construct S graphs from the event stream (where $S = 1$ for object classification), and each graph is passed through a spatial feature learning module. For object classification, the output of this module is mapped to object classes directly by fully connected layers. For action recognition and action similarity labeling, we model coarse temporal dependencies over multiple graphs by converting to a
grid representation via the Graph2Grid module and perform temporal feature learning with a conventional 3D CNN.}
        \label{f:framework}
\end{figure*}

The architecture of our graph-based spatio-temporal feature learning network (Fig. \ref{f:framework}) and comprises four parts: sampling and graph construction, a spatial feature learning module, a graph-to-frame mapping module and a temporal feature learning module.  For object classification, a single graph is constructed, whereas for action-based tasks with longer temporal extent, multiple graphs are extracted over the event stream duration. Specifically, neuromorphic events are firstly sampled and  represented by a sequence of graphs. Graphs are then individually processed by a spatial feature learning module, which consists of multiple graph convolution and pooling layers to map the input to a coarser graph encoding.  For object classification, we obtain a single graph encoding that we pass to a single fully connected layer for prediction. Conversely, for action recognition and action similarity labeling, we obtain multiple graph encodings. As such, we convert the graphs to a grid representation with a graph-to-frame mapping module which we denote as Graph2Grid, and stack the resulting frames for temporal feature learning with a 3D CNN.  In this way, we are able to effectively and efficiently learn spatio-temporal features for motion-based applications, such as action recognition. We provide more details on each component of the framework in the following sections.
%We provide more details on each component of the framework in the following sections.

\subsection{Graph Construction} \label{sec:graph_construction}
Given a NVS sensor with spatial address resolution of $H \times W$, we express a volume of events $V$ produced by a NVS camera as a tuple sequence: 
\begin{equation}
\{e_{i}\}_{N} = \{x_{i},y_{i},t_{i},p_{i}\}_{N}
\end{equation}
where $(x_{i},y_{i})\in \{1,2, \dots H\}\times\{1,2, \dots W\}$ is the spatial address at which the spike event occurred, $t_{i}$ is the timestamp indicating when the event was generated,  $p_{i} \in{{\{+1,-1\}}}$ is the event polarity (with +1/-1 signifying ON/OFF events respectively), and $N$ is the total number of events. 

 To reduce the storage and computational cost, we use  non-uniform grid sampling  \cite{lee2001point} to sample a subset of $M$  representative events $\{e_{i}\}_{M} \subset\{e_{i}\}_{N}$, where $M \ll\ N$. Effectively, one event is randomly selected from a space-time volume with the maximum number of events inside. If we consider $\mathbf{s}\{e_{i}\}_{i=1}^{k}$ to be such a volume containing $k$ events, then only one event $e_{i}$ ($i \in [1,k]$) is randomly sampled in this space-time volume.
We then define the sampled events $\{e_{i}\}_{\{M\}}$ on a directed graph $\mathrm{G}=\{\mathrm{\nu},\mathrm{\varepsilon},\mathrm{U}\}$, with $\mathrm{\nu}$ being the set of vertices, $\mathrm{\varepsilon}$ the set of the edges, and $\mathrm{U}$ the coordinates of the nodes that locally define the spatial relations of the nodes. The sampled events are independent and not linked, therefore, we regard each event $e_{i} : (x_{i},y_{i},t_{i},p_{i})$ as a node in the graph, such that $\nu_{i}:(x_{i},y_{i},t_{i})$, with $\nu_{i} \in \mathrm{\nu}$. We define the connectivity of nodes in the graph based on the radius-neighborhood-graph strategy. Namely, nodes $\nu_{i}$ and $\nu_{j}$ are connected with an edge only if their weighted Euclidean distance $d_{i,j}$ is less than radius distance $R$. For two spike events  $e_{i}$ and $e_{j}$, the Euclidean distance between them is defined as the weighted spatio-temporal distance: 
\begin{equation}
d_{i,j}=\sqrt{\alpha(|x_{i}-x_{j}|^{2}+|y_{i}-y_{j}|^{2})+\beta|t_{i}-t_{j}|^{2}}\leq \mathrm{R}
\end{equation}
where $\alpha$ and $\beta$ are weight parameters compensating for the difference in spatial and temporal grid resolution (timing accuracy is significantly higher in NVS cameras than spatial grid resolution). To limit the size of the graph, we constrain the maximum connectivity degree for each node by parameter $D_{\max}$. We subsequently define $u(i,j)$  for node  $i$, with connected node  $j$, as $u(i,j)=\left [ \left | x_{i}-x_{j} \right |, \left | y_{i}-y_{j} \right |\right ] \in \mathrm{U}$. 

After connecting all nodes of the graph $\mathrm{G}=\{\mathrm{\nu},\mathrm{\varepsilon}, \mathrm{U}\}$ via the above process, we consider the polarity of events as a signal that resides on the nodes of the graph $\mathrm{G}$. In other words, we define the input feature for each node  $i$, as $f^{(0)}(i)=p_i \in \{+1,-1\}$.

We introduce the parameter $S$ to represent the number of graphs constructed from one sample. Given that object classification is appearance-based and typically only requires a short temporal extent, we set $S=1$. Specifically, we randomly extract $T_{\mathrm{vol}}$ length events over the entire event stream to construct a graph. Conversely, for action recognition and action similarity labeling, we divide the event stream into $S$ volumes with the same time duration $T/S$, where $T$ is the sample duration. We then construct a graph for each volume in which $T_{\mathrm{vol}}<T/S$ length events are randomly extracted to construct a graph, giving us a set of graphs $\mathcal{G}=\{\mathrm{G}_n\}_{n=1}^{S}$. In this way, we efficiently model coarse temporal dependencies over the duration of the sample, without constructing a  single large and substantially complex graph. The graphs can thus be processed individually by our spatial feature learning module before fusion with our Graph2Grid module and temporal feature learning.  This is inspired by recent work on  action recognition with RGB frames \cite{feichtenhofer2016convolutional}, which fuses representations over coarse temporal scales with 3D convolutions and pooling; indeed, our graph-based framework is substantially more lightweight and does not suffer from the limitations of active pixel sensing.

\subsection{Spatial Feature Learning Module}\label{sec:spatiallearning}
The constructed graphs are first fed individually into a spatial feature learning module,  where our framework learns appearance information.  According to the common architectural pattern for feed-forward neural networks, these graph convolutional neural networks are built by interlacing graph convolution layer and graph pooling layers, where the graph convolution layer performs a non-linear mapping and the pooling layer reduces the size of the graph.  

Graph convolution generalizes the convolutionl operator to the graph domain. Similar to frame-based convolution,  graph convolution  can be categorized into two types: spectral and spatial. Spectral convolution  \cite{defferrard2016convolutional, bruna2013spectral} defines the convolution operator by decomposing a graph in the spectral domain and then applying a spectral filter on the spectral components. However, this operation requires identical graph input and handles the whole graph simultaneously, so it is not suitable for  the variable and large graphs constructed from NVS. On the other hand, spatial convolution  \cite{fey2018splinecnn, masci2015geodesic} aggregates a new feature vector for each vertex, using its neighborhood information weighted by a trainable kernel function. Because of this property, we consider  spatial convolution operation as a better choice when dealing with graphs from NVS. 

Similar to conventional frame-based convolution, spatial convolution operations on graphs are also a one-to-one mapping between kernel function and neighbors at relative positions w.r.t. the central node of the convolution. Let $i$ denote a node of the graph with feature $f(i)$, $\mathcal{N}(i)$ denote the set of neighbors of node $i$ and  $g(u(i,j))$ denote the weight parameter constructed from the kernel function $g(\cdot)$. The graph convolution operator $\otimes$ for this node can then be written in the following general form: 
\begin{equation} \label{eq:graph_conv}
(f \otimes g)(i) = \frac{1}{\left | \mathcal{N}(i) \right |}\sum_{j\in\mathcal{N}(i)}f(j)\cdot g(u(i,j))
\end{equation}
where $\left | \mathcal{N}(i) \right |$ is the cardinality of $\mathcal{N}(i)$. We can generalize (\ref{eq:graph_conv}) to  multiple input features per node. Given the kernel function $\mathbf{g}=(g_{1},...,g_{l},...,g_{M_{in}})$ and input node feature vector $\mathbf{f}_l$, with $M_{in}$ feature maps indexed by $l$, the spatial convolution operation $\otimes$ for the node $i$ with $M_{in}$ feature maps is defined as:
\begin{equation}
(\mathbf{f}\otimes\mathbf{g})(i) = \frac{1}{\left | \mathcal{N}(i) \right |} \sum_{\l=1}^{M_{in}}\sum_{j\in\mathcal{N}(i)}f_{l}(j)\cdot g_{l}(u(i,j))
\label{eq:CONV}
\end{equation}

The kernel function $\mathbf{g}$ defines how to model the coordinates $\mathrm{U}$. The content of  $\mathrm{U}$ is used to determine how the features are aggregated and the content of $f_{l}(j)$ defines what is aggregated. As such, several spatial convolution operations \cite{fey2018splinecnn, masci2015geodesic, monti2017geometric} on graphs were proposed by using different choice of kernel functions. Among them, SplineCNN \cite{fey2018splinecnn} achieves state-of-the-art results in several applications, so in our work we use the same kernel function as in SplineCNN. In this way, we leverage properties of B-spline bases to efficiently filter NVS\ graph inputs of arbitrary dimensionality. Let $((N_{1,i}^{m})_{1\leq  i\leq k_{1}},...,(N_{d,i}^{m})_{1\leq  i\leq k_{d}})$ denote $d$ open $\mathbf{B}$-spline bases of degree $m$ with $\mathbf{k}=(k_{1},...,k_{d})$ defining $d$-dimensional kernel size \cite{piegl2012nurbs}. Let $w_{\boldsymbol{z},l} \in \mathbf{W}$ denote a trainable parameter for each element $\boldsymbol{z}$ from the Cartesian product $\mathcal{Z}= (N_{1,i}^{m})_{i}\times\cdot \cdot \cdot \times (N_{d,i}^{m})_{i}$ of the B-spline bases and each of the $M_{in}$ input feature maps indexed by $l$. Then the kernel function $g_{l}:[a_{1},b_{1}]\times\cdot \cdot \cdot \times[a_{d},b_{d}]\rightarrow \mathbb{R}$ is defined as
\begin{equation}
g_{l}(\mathbf{u})=\sum_{\boldsymbol{z}\in \mathcal{Z}}w_{\boldsymbol{z},l} \cdot \prod _{s=1}^{d}N_{s,z_{s}}(u_{s})
\end{equation}
We denote a graph convolution layer as $\mathrm{Conv}(M_{\mathrm{in}},M_{\mathrm{out}})$, where $M_{\mathrm{in}}$ is the number of input feature maps and $M_{\mathrm{out}}$ is the number of output feature maps indexed by $l^{'}$. Then, a graph convolution layer with bias $b_{l}$ and activation function $\xi (t)$, can be written as:
\begin{eqnarray}\label{eq:conv_layer}
\mathrm{Conv}_{l^{'}}  &=&  \xi \Big{( }\frac{1}{\left | \mathcal{N}(i) \right |} \sum_{\l=1}^{M_{\mathrm{in}}}\sum_{j\in\mathcal{N}(i)}f_{l}(j)\cdot \sum_{\boldsymbol{z} \in \mathcal{Z}}w_{z,l}  \\ & \cdot &  \prod _{s=1}^{d}N_{s,z_{s}}(u_{s}) + b_{l^{'}}\Big{)} \nonumber  
\end{eqnarray}
where $l^{'}=1,..,M_{\mathrm{out}}$, indicates the $l^{'}$th output feature map. This defines a single graph convolutional layer.  For  $C$ consecutive graph convolutional layers, $(\mathrm{Conv}^{(c)})_{c\in[0,C]}$, the $c$-th layer has a corresponding input feature map $\mathbf{f}^{(c)}$ over all nodes, with the input feature for node $i$ of the first layer  $\mathrm{Conv}^{(0)}$,  $f^{(0)}(i)=p_i\in{{\{+1,-1\}}}$.   

To accelerate deep network training, we use batch normalization \cite{ioffe2015batch} before the activation function. That is, the whole node feature $f_{l^{'}}$ over the $l^{'}$-th channel map is normalized individually via
\begin{equation}\label{eq:batch_norm}
f^{'}_{l^{'}}= \frac{f_{l}-E(f_{l^{'}})}{\sqrt{\mathrm{Var}(f_{l^{'}})+\epsilon }}\cdot \gamma +\beta \quad l^{'}=1,..,M_{\mathrm{out}}
\end{equation}
where $E(f_{l^{'}})$ and $\mathrm{Var}(f_{l^{'}})$ denote mean and variance of $f_{l^{'}}$ respectively, $\epsilon$ is used to ensure normalization does not overflow when the variance is near zero, and $\gamma$ and $\beta$ represent trainable parameters.

\textbf{Residual Graph CNNs}: Inspired by the  ResNet architecture  \cite{he2016deep}, we propose  residual graph CNNs for our spatial feature learning module, in order  to resolve the well-known degradation problem inherent with increasing number of layers (depth) in graph CNNs \cite{li2018deeper}. Our residual graph CNN (RG-CNN) is effectively composed of a  series of residual blocks and pooling layers. Considering equations (\ref{eq:conv_layer}) and (\ref{eq:batch_norm}) denote a single graph convolutional layer with batch normalization \cite{ioffe2015batch} that accelerates the convergence of the learning process, we apply residual connections in spatial feature learning module by summing element-wise the outputs of graph convolutions. Our ``shortcut'' connection comprises a graph convolution layer with kernel size $K=1$ for mapping the feature dimension to the correct size, and is also followed by batch normalization. A residual block is illustrated at the bottom left of Fig. \ref{f:framework}. We denote the resulting  graph residual block as  $\mathrm{Res_{g}(c_{in},c_{out})}$, with $c_{\mathrm{in}}$ input feature maps and $c_{\mathrm{out}}$ output feature maps.

A residual block is followed by max pooling over clusters of nodes;
given a graph representation, let us denote the  spatial coordinates for node $i$ as $(x'_{i},y'_{i})\in \{1,2,\dots H'\} \times \{1,2,\dots W'\} $ and resolution as  $H' \times W'$. We define the cluster size as $s_h \times s_w$, which corresponds to the downscaling factor in the pooling layer of   $\left \lceil \frac{H'}{s_h} \right \rceil \times \left \lceil \frac{W'}{s_w} \right \rceil$.  For each cluster, we generate a single node, with feature  set to  the  maximum over node features $\mathbf{f}$ in the cluster, and coordinates set  to the average of node coordinates $(x'_{i},y'_{i})$ in the cluster. 
%\begin{equation}
%\begin{equation}
%\left\{\begin{matrix}
%x_{\mathrm{new}}= \left \lfloor \sum_{i=1}^{\mathrm{num}}x_{i}/\mathrm{num}  \right \rfloor      \\ 
%y_{\mathrm{new}}= \left \lfloor \sum_{i=1}^{\mathrm{num}}y_{i}/\mathrm{num}  \right \rfloor.
%\end{matrix}\right.
%\end{equation}
Importantly, if there are connected nodes between two clusters, we assume the new generated nodes in these two clusters are connected with an edge. % $s_h$ and $s_w$ are the dimension size of clusters, allowing for various down-scaling. 

 For object classification, where the entire event stream can be modelled by a single graph, we can directly map the output of the spatial feature learning module to the classes with a fully connected layer.   Given $M_{\mathrm{in}}$ feature maps $\mathbf{f}\in\mathbb{R}^{I\times M_{\mathrm{in}}}$ from a graph with $I$ nodes, similar to CNNs, a fully connected layer in a graph convolutional network is a weighted linear combination linking all input features to outputs. Let us denote ${f}^\mathrm{spatial}_{l}(i)$ as the $l$th output feature map of the $i$th node of the spatial feature learning module, then we can derive a fully connected layer in the graph as:
\begin{equation}
\label{eq:fout_q}f_{q}^{\mathrm{FC}}=\xi \Big{(}\sum_{i=1}^{I}\sum_{l=1}^{M_{\mathrm{in}}}F_{i,l,q}{f}^\mathrm{spatial}_{l}(i)\Big{)} \quad q=1,...,Q
\end{equation}
where $Q$ is the number of output channels indexed by $q$, $F$ is an array of trainable weights with size ${I \times M_{\mathrm{in}} \times Q}$, $\xi (t)$ is the non-linear activation function, e.g. ReLU: $\xi (t) = \max{(0,t)}$. For the remainder of the paper, we use  $\mathrm{FC}(Q)$ to indicate a fully connected layer with $Q$ output dimensions.

\subsection{Graph2Grid: From Graphs to Grid Snippet}
For motion-based tasks, we need to model temporal dependencies over the entire event stream.  As discussed in Section \ref{sec:graph_construction}, given a long sample duration, it is not feasible to construct  a single graph over the entire event stream, due to the sheer number of events.  It is more computationally feasible to generate multiple graphs for time blocks of duration $T_{\mathrm{vol}}$.  These are processed individually by the  spatial feature learning module. However, to model coarse temporal dependencies over multiple graphs, we must fuse the spatial feature representations.  We propose a new Graph2Grid module that transforms the learned graphs from our spatial feature learning module to a grid representation and performs stacking over temporal dimension, as illustrated in Fig. \ref{f:framework}. In this way, we are effectively able to create pseudo frames from the graphs, with $M_{in}$ channels   and timestamp $(n-1)T_{\mathrm{vol}}$, corresponding to the $n$-th graph.

Again, denoting the output spatial feature learning map as ${f}^\mathrm{spatial}_{l}(i)$ for the $l$th output feature map of the $i$th node with coordinates $(x'_{i},y'_{i})\in \{1,2,\dots, H_\mathrm{spatial}\} \times \{1,2,\dots, W_\mathrm{spatial}\}$, we define a grid representation $\mathbf{f}^\mathrm{grid}$ of spatial size  $H_\mathrm{spatial} \times W_\mathrm{spatial}$ as follows:    
  \begin{equation}
    f^{\mathrm{grid}}_{a,b,l}=
    \begin{cases}
      f^\mathrm{spatial}_{l}(i), & \text{when}\ a=x_i', b=y_i'  \\
      0, & \text{otherwise}
    \end{cases}
  \end{equation}
where $(a,b) \in \{1,2,\dots, H_\mathrm{spatial}\} \times \{1,2,\dots, W_\mathrm{spatial}\}$. The resulting grid feature representation $\mathbf{f}^\mathrm{grid} \in \mathbb{R}^{H_\mathrm{spatial} \times W_\mathrm{spatial} \times M_{in}}$ is for
a single graph; for $S$ graphs over the temporal sequence, we simply concatenate over a fourth temporal dimension. \ We denote the resulting grid feature over $S$ graphs as $\mathbf{F}^\mathrm{grid} = \mathbf{f}^{\mathrm{grid},1}||\mathbf{f}^{\mathrm{grid},2}||\dots ||\mathbf{f}^{\mathrm{grid},S}$, where $||$ denotes concatenation over the temporal axis.  Thus, the dimensions of $\mathbf{F}^\mathrm{grid} $ are $H_\mathrm{spatial} \times W_\mathrm{spatial} \times M_\mathrm{in} \times S$.  This grid feature matrix can therefore be fed to a conventional 3D convolutional neural network in our temporal feature learning module, in order to learn both the coarse temporal dependencies, but also a full spatio-temporal representation of the input.   

\subsection{Temporal Feature Learning Module} \label{sec:temporallearning}
The output feature matrix  $\mathbf{F}^\mathrm{grid}$ contains both spatial and temporal information over the entire sample duration, which can be effectively encoded  with a conventional 3D CNN  \cite{tran2015learning} in order to generate a final spatio-temporal representation of the video input for action recognition. In this paper, we consider three network architectures for the 3D\ CNN;  a plain architecture with interlaced 3D convolutional and pooling layers, an I3D-based architecture  comprising multiple I3D blocks as configured in \cite{carreira2017quo}, and a 3D residual block design. Our 3D residual block design  is illustrated in the bottom right of Fig. \ref{f:framework}; essentially for  $C$ consecutive convolutional layers, every $(c-2)$-th  layer is connected to the  $c$-th layer via a non-linear residual connection, for all $c\in\{3,5\dots C-2, C\}$,  and every layer is followed by batch normalization.   For all architectures, we aggregate the features in the final layer of the CNN with global average pooling and pass to a fully connected layer for classification. We provide further experimental details in Section \ref{sec:applications}, describing the number of input and  output channels per layer.

It is worth noting that while 3D CNNs are notorious for being computationally heavy, typical NVS cameras like the iniLabs DAVIS240c has spatial resolutions of the order of $240 \times 180$; in conjunction with the use of pooling in our spatial feature learning module, this means that the spatial size of $\mathbf{F}^\mathrm{grid}$ is at most $30 \times 30$. This is substantially lower input resolution than APS-based counterparts ingesting RGB frames, where the spatial resolution to the 3D CNN is typically $224 \times 224 $ or higher. \    \

\section{Experimental Details and Evaluation}\label{sec:applications}
In this section, we demonstrate the potential of our framework as a method of representation learning for high-level computer vision tasks with NVS inputs. In Section \ref{sec:objectclassification}, we focus on object classification as an appearance-based application. Then in Sections \ref{sec:recognition} and \ref{sec:similarity}, we present results for large-scale multi-class human action recognition and action similarity labeling as motion-based applications.
Beyond evaluation on standard datasets, we introduce our newly proposed ASL-DVS\ dataset in Section \ref{sec:objectclassification}, which is the largest-source dataset for object classification. We additionally generate the largest NVS-based action recognition and action similarity labelling datasets by converting standard APS datasets, UCF101, HMDB51 and ASLAN, to the NVS domain and explain the recording process prior to evaluation in Sections \ref{sec:recognition} and \ref{sec:similarity} respectively.

\subsection{Object Classification}\label{sec:objectclassification}
\begin{figure}
\vspace{-0.05in}
  \centering
  \includegraphics[width=3.3in]{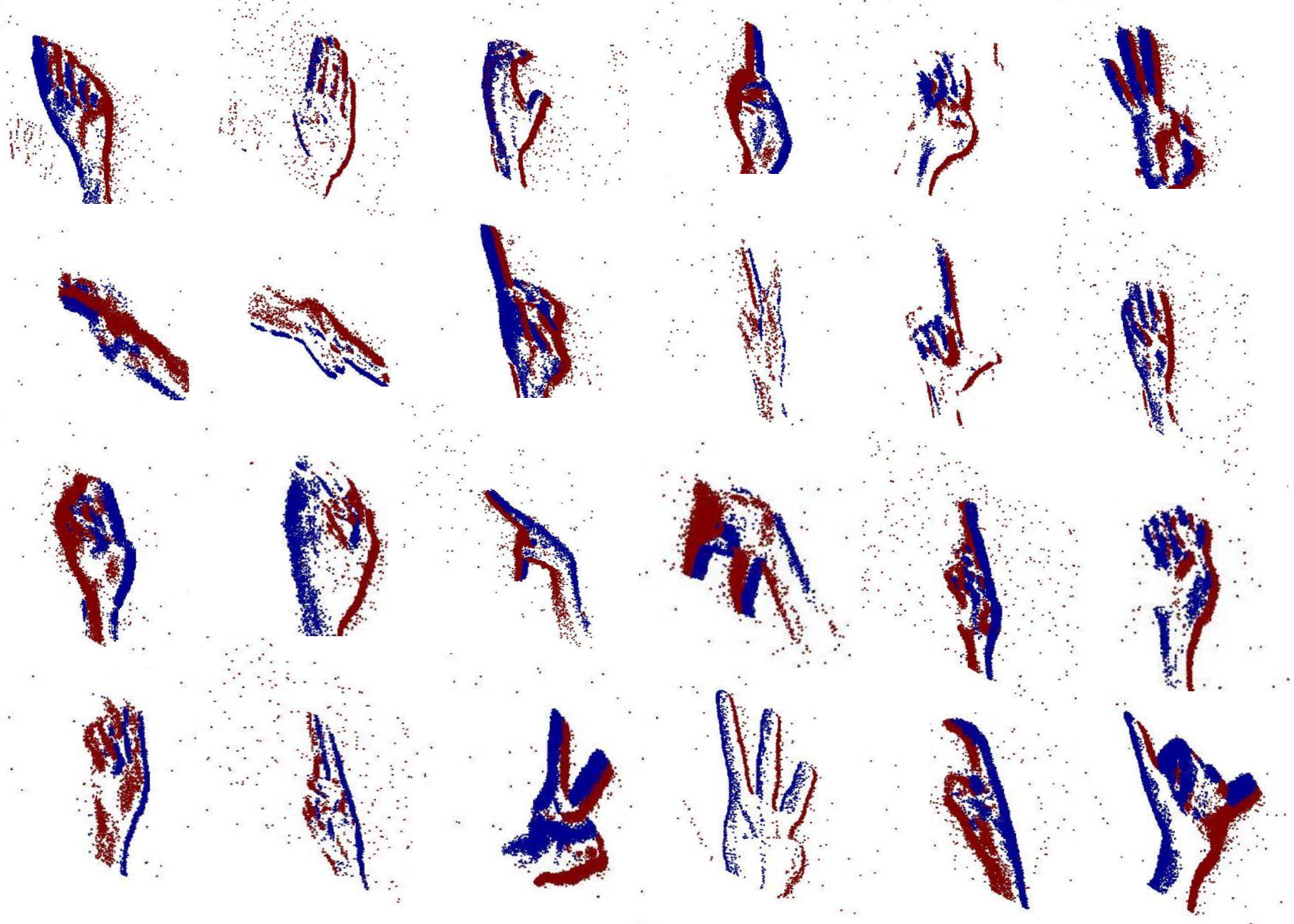}
  \vspace{-0.05in}
  \caption{Examples of the ASL-DVS dataset (the visualizations correspond to letters A-Y, excluding J, since letters  J and Z involve motion rather than static shape). Events are grouped to image form for visualization (Red/Blue: ON/OFF events).}
        \label{f:ASL}
\end{figure}

\begin{figure}
\vspace{-0.15in}
  \centering
  \includegraphics[width=3.2in]{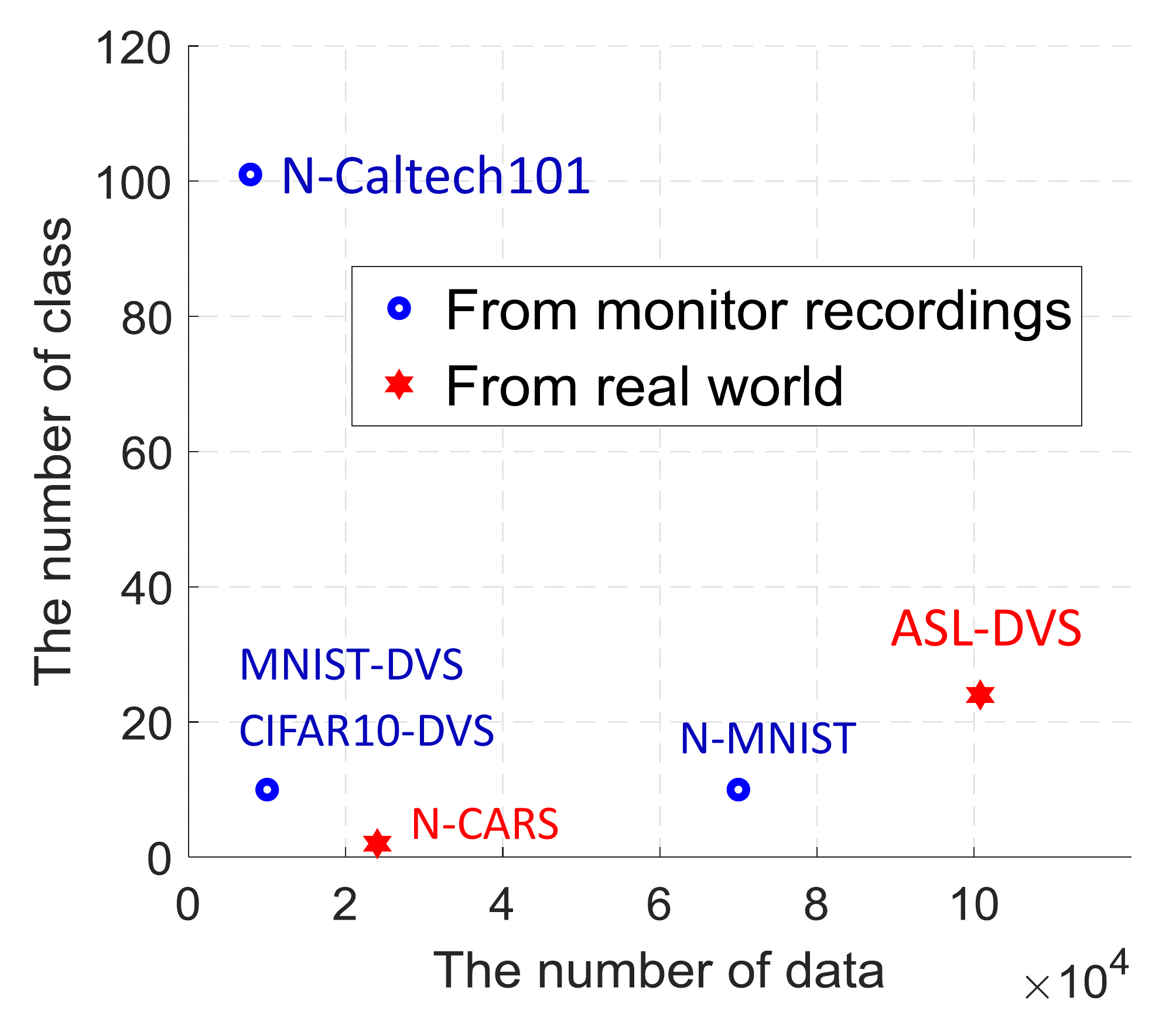}
  \vspace{-0.1in}
  \caption{Comparison of NVS datasets w.r.t. the number of classes and the total size. }
                \label{f:Datasets}
\vspace{-0.15in}
\end{figure}

%Object classification finds numerous applications in visual surveillance, human-machine interfaces, image retrieval and visual content analysis systems. We first introduce the datasets we evaluate on, including our new ASL-DVS dataset, before discussing implementation details and presenting results. We compare with recent state-of-the-art methods and perform complexity analysis.  

\textbf{Datasets:} Many neuromorphic datasets for object classification are converted from standard frame-based datasets, such as N-MNIST \cite{orchard2015converting}, N-Caltech101 \cite{orchard2015converting}, MNIST-DVS \cite{serrano2015poker} and CIFAR10-DVS \cite{li2017cifar10}. N-MNIST and N-Caltech101 were acquired by an ATIS sensor \cite{posch2011qvga} moving in front of an LCD monitor while the monitor is displaying each sample image. Similarly, MNIST-DVS and CIFAR10-DVS datasets were created by displaying a moving image on a monitor and recording with a fixed DAVIS sensor \cite{lichtsteiner2008128}. Emulator software has also been proposed in order  to generate neuromorphic events from pixel-domain video formats using the change of pixel intensities of successively rendered images \cite{mueggler2017event, bi2017pix2nvs}. While useful for early-stage evaluation, these datasets  cannot capture the real dynamics of an NVS\ device due to the limited frame rate of the utilized  content, as well as the limitations and artificial noise imposed by the recording or emulation environment. To overcome these limitations, N-CARS dataset \cite{sironi2018hats} was created by directly recording objects in urban environments with an ATIS sensor. 
%This  two-class real-world dataset comprises 12,336 car samples and 11,693 non-car samples (background) with 0.1 second length. 
Despite its size, given that it only corresponds to a binary classification problem, N-CARS cannot represent the behaviour of object classification algorithms on more complex NVS-based tasks. 

We present a large 24-class dataset  of handshape recordings under realistic conditions. Its 24 classes correspond to 24 letters (A-Y, excluding J) from the American Sign Language (ASL), which we call ASL-DVS. Examples of recordings  are shown in Fig \ref{f:ASL}. The ASL-DVS was recorded with an iniLabs DAVIS240c NVS camera set up in an office environment with low ambient noise and constant illumination. For all recordings, the camera was at the same position and orientation to the persons carrying out the handshapes. Five subjects were asked to pose the different static handshapes relative to the camera in order to introduce natural variance into the dataset. For each letter, we collected 4,200 samples (total of 100,800 samples) and each sample lasts for approximately 100 milliseconds. Fig. \ref{f:Datasets} shows a comparison of existing NVS\ datasets w.r.t. the number of classes and the total size. Within the landscape of existing datasets, our ASL-DVS is a comparably complex dataset with the largest number of labelled examples. We, therefore, hope that this will make it a useful resource for researchers to build comprehensive models for NVS-based object recognition, especially given the fact that it comprises real-world recordings. ASL-DVS will be publicly available for download at a link to be provided after the review process is completed. 
%In our experiments, all datasets presented  in Fig. \ref{f:Datasets}  are used to validate our algorithm. 

\begin{table*}[ht]
\centering
\caption{Top-1 accuracy of our graph CNNs w.r.t. the state-of-the-art, other graph convolution networks and deep CNNs. }
\vspace{-0.1in}
\label{t:toART}
\begin{tabular}{cccccccc}
\hline
Model & N-MNIST & MNIST-DVS & N-Caltech101 & CIFAR10-DVS & N-CARS & ASL-DVS\\
\hline
H-First \cite{orchard2015hfirst} & 0.712 & 0.595 & 0.054 & 0.077 & 0.561 &-\\
HOTS \cite{lagorce2017hots} & 0.808 & 0.803 & 0.210 & 0.271 &0.624 &-\\
Gabor-SNN \cite{lee2016training,neil2016phased} & 0.837 & 0.824 & 0.196 & 0.245 & 0.789 &-\\
HATS \cite{sironi2018hats} & \textbf{0.991} & 0.984 & 0.642 & 0.524 & 0.902 &-\\
\hline
GIN \cite{xu2018powerful} & 0.754 & 0.719 & 0.476 & 0.423 & 0.846 & 0.514\\
ChebConv \cite{defferrard2016convolutional} & 0.949 & 0.935 & 0.524 & 0.452 & 0.855 & 0.317\\
GCN \cite{kipf2016semi} & 0.781 & 0.737 & 0.530 & 0.418 & 0.827 & 0.811\\
MoNet \cite{monti2017geometric} & 0.965 & 0.976 & 0.571 & 0.476 & 0.854 & 0.867\\
\hline
VGG\_19 \cite{simonyan2014very} & 0.972 & 0.983 & 0.549 & 0.334 & 0.728 & 0.806\\
Inception\_V4 \cite{szegedy2017inception} & 0.973 & 0.985 &0.578 & 0.379 & 0.864 & 0.832\\
ResNet\_50 \cite{he2016deep} & 0.984 & 0.982 & 0.637 &  \textbf{0.558} & 0.903 & 0.886\\
\hline
G-CNNs  & 0.985 & 0.974 & 0.630 & 0.515 & 0.902& 0.875\\
RG-CNNs (proposed) & 0.990 & \textbf{0.986} & \textbf{0.657} & {0.540} & \textbf{0.914} &\textbf{0.901}\\
\hline             
\end{tabular}
\end{table*}

\textbf{Implementation Details:} For simple datasets N-MNIST and MNIST-DVS, our spatial feature learning module is only comprised of two graph residual blocks. Graph residual blocks are described in Section \ref{sec:spatiallearning}, and we fix the kernel size $K=5$  for all convolutional layers outside of the skip connection. We denote a graph convolutional layer as $\mathrm{Conv}_\mathrm{g}(c_\mathrm{in},c_\mathrm{out}), $ fully connected layer as $\mathrm{FC}(c_\mathrm{in},c_\mathrm{out})$ and graph residual block as $\mathrm{Res_{g}(c_\mathrm{in},c_\mathrm{out}),  }$ where $c_\mathrm{in}$ and $c_\mathrm{out}$ are the input and output channels respectively. Additionally, we denote max graph pooling layers as $\mathrm{MaxP_\mathrm{g}(s_\mathrm{h}, s_\mathrm{w}),}$ where $s_\mathrm{h}$ and $s_\mathrm{w}$ represent the cluster size.  
With this notation, the architecture of our network for these can be written as  $\mathrm{Conv_g(1,32)}$$\longrightarrow$$\mathrm{MaxP_g(2,2)}$$\longrightarrow$$\mathrm{Res_{g}(32,64)}$ $\longrightarrow$$\mathrm{MaxP_g(4,4)}$$\longrightarrow$$\mathrm{Res_{g}(64,128)}$$\longrightarrow$$\mathrm{MaxP_g(7,7)}$$\longrightarrow$ $\mathrm{FC(128,128)}$$\longrightarrow$$\mathrm{FC}(128,Q)$, where $Q$ is the number of classes of each dataset. 
For the remaining datasets, three residual graph blocks  are used, and the utilized network architecture is $\mathrm{Conv_g(1,64)}$$\longrightarrow$$\mathrm{MaxP_g(s_\mathrm{h}, s_\mathrm{w})}$$\longrightarrow$$\mathrm{Res_{g}(64,128)}$$\longrightarrow$ $\mathrm{MaxP_g(s_\mathrm{h}, s_\mathrm{w})}$$\longrightarrow$$\mathrm{Res_{g}(128,256)}$$\longrightarrow$$\mathrm{MaxP_g(s_\mathrm{h}, s_\mathrm{w})}$$\longrightarrow$ $\mathrm{Res_{g}(256,512)}$$\longrightarrow$$\mathrm{MaxP_g(s_\mathrm{h}, s_\mathrm{w})}$$\longrightarrow$$\mathrm{FC(512,1024)}$$\longrightarrow$\\$\mathrm{FC}(1024,Q)$. Since the datasets are recorded from different sensors, the spatial resolution of each sensor is different (i.e., DAVIS240c: 240$\times$180, DAVIS128 \& ATIS: 128$\times$128), leading to various maximum coordinates for the graph.  We, therefore, set the cluster size in pooling layers in two categories; \textit{(i)} N-Caltech101 and ASL-DVS: 4$\times$3, 16$\times$12, 30$\times$23 and 60$\times$45; \textit{(ii)}\  CIFAR10-DVS and N-CARS: 4$\times$4, 6$\times$6, 20$\times$20 and 32$\times$32. 
We also compare the proposed residual graph networks (RG-CNNs) with their corresponding plain graph networks (G-CNNs), which utilize the same number of graph convolutional and pooling layers but without the residual connections. The degree of B-spline bases $m$ of all convolutions in this work is set to 1.

\textbf{}For the N-MNIST, MNIST-DVS and N-CARS datasets, we use the predefined training and testing splits, while for N-Caltech101, CIFAR10-DVS and ASL-DVS, we follow the experiment setup of Sironi  \cite{sironi2018hats}: 20\% of the data is randomly selected for testing and the remaining is used for training. For each sample, events within 30-millisecond length are randomly extracted to input to our object classification framework. During the non-uniform sampling, the maximal number of events $k$ in each space-time volume is set to 8. When constructing graphs, the radius $R$ is 3, weight parameters $\alpha$ and $\beta$ are set to 1 and $0.5 \times 10^{-5}$, respectively, the maximal connectivity degree $D_{\max}$ for each node is 32, and $T_\mathrm{vol}=1/30s$ length events are randomly extracted to form the graph.
In order to reduce overfitting, we add dropout with probability 0.5 after the first fully connected layer and also perform data augmentation. In particular, we spatially scale node positions by a randomly sampled factor within $[0.95,1)$, perform mirroring (randomly flip node positions along 0 and 1 axis with 0.5 probability) and rotate node positions around a specific axis by a randomly sampled factor within $[0,10]$ in each dimension. Networks are trained with the Adam optimizer and the cross-entropy loss between softmax output and the one-hot label distribution for 150 epochs with batch size 64 and learning rate 0.001 step-wise decreasing by 0.1 after 60 and 110 epochs.

\textbf{Results:} We compare Top-1 classification accuracy obtained from our model with that from HOTS \cite{lagorce2017hots}, H-First \cite{orchard2015hfirst}, SNN \cite{lee2016training,neil2016phased} and HATS \cite{sironi2018hats}. We report  results from Sironi \textit{et al.}  \cite{sironi2018hats}, since we use the same training and testing methodology. The results are shown in Table \ref{t:toART}. For the simple  N-MNIST and MNIST-DVS datasets, whose accuracy is already close to near-perfect classification, our models achieve  comparable results. For the other datasets,  our proposed RG-CNNs consistently set the new state-of-the-art on these datasets. 

Table \ref{t:toART} also includes the classification results stemming from other graph convolutional networks; namely, GIN  \cite{xu2018powerful}, ChebConv  \cite{xu2018powerful}, GCN \cite{kipf2016semi} and MoNet \cite{monti2017geometric}. The architectures of these networks are the same as our plain graph networks (G-CNNs) introduced in this section, with the only difference being the graph convolutional operation. The training details and data augmentation methods are the same as illustrated before.  The Top-1 classification accuracy stemming from all networks of Table \ref{t:toART} indicates that our proposed RG-CNN and G-CNN outperform all the other graph convolutional networks.

To further validate our proposal, we compare our results with conventional deep convolutional networks. There are no conventional CNNs specifically designed for NVS events, so we train/evaluate on three well-established CNNs, namely VGG\_19 \cite{simonyan2014very}, Inception\_V4 \cite{szegedy2017inception} and ResNet\_50 \cite{he2016deep}. The format of the required input for these CNNs is frame-based, so we convert neuromorphic spike events to frame form similarly to the grouping of Zhu \textit{et al.} \cite{zhu2018ev}. We thereby introduce a two-channel event image form with the same resolution as the NVS sensor: the two channels encode the number of positive and negative events that have occurred at each position. In addition, each frame grouping corresponds to  a random time segment of 30 ms of spike events. To avoid overfitting, we supplement the training with heavy data augmentation: first, we resize the input images such that the smaller dimension is 256 and keep the aspect ratio; then, we use a random cropping of 224$\times$224 spatial samples of the resized frame; finally, the cropped volume is  randomly flipped and normalized according to its mean and standard deviation. We train all CNNs from scratch using stochastic gradient descent with momentum set to 0.9 and $L_{2}$ regularization set to $0.1 \times 10^{-4}$. The learning rate is initialized at $10^{-3}$ and decayed by a factor of 0.1 every 10k iterations.  As shown in Table \ref{t:toART}, despite performing comprehensive data augmentation and $L_{2}$ regularization to avoid overfitting, the results acquired from conventional CNNs are still below the-state-of-the-art since event images contain far less information (see Fig. \ref{f:example}). Thus, except for the CIFAR10-DVS dataset, the accuracy of our proposals surpasses that of conventional frame-based deep CNNs. 

\begin{table}
\caption{Complexity (GFLOPs) and size (MB) of networks.}
\vspace{-0.1in}
\label{t:flops}
\centering
\begin{tabular}{ccc}
\hline
Model & GFLOPs & Size (MB) \\
\hline
VGG\_19 \cite{simonyan2014very} & 19.63 & 143.65\\
Inception\_V4 \cite{szegedy2017inception} & 12.25 & 42.62 \\
ResNet\_50 \cite{he2016deep} & 3.87 & 25.61\\
\hline
G-CNNs & 0.39 & 18.81\\
RG-CNNs & 0.79 & 19.46\\
\hline             
\end{tabular}
\vspace{-0.1in}
\end{table}

\textbf{Complexity Analysis:} We now turn our attention to the complexity of our proposals and compare the number of floating-point operations (FLOPs) and the number of parameters of each model. In conventional CNNs, we compute FLOPs for convolution layers as \cite{molchanov2016pruning}:
\begin{equation}
\mathrm{FLOPs} = 2HW(C_{\mathrm{in}}K^{2}+1)C_{\mathrm{out}}
\end{equation}
where $H$, $W$ and $C_{\mathrm{in}}$ are height, width and the number of channels of the input feature map, $K$ is the kernel size, and $C_{\mathrm{out}}$ is the number of output channels. For graph convolution layers, FLOPs stem  from 3 parts  \cite{fey2018splinecnn}; \textit{(i)}\ for computation of B-spline bases, there are $N_{\mathrm{edge}}(m+1)^{d}$ threads each performing $7d$ FLOPs (4 additions and 3 multiplications), where $N_{\mathrm{edge}}$ is the number of edges, $m$ the B-spline basis degree and $d$ the dimension of graph coordinates; \textit{(ii)} for convolutional operations, the FLOPs count is $3N_{\mathrm{edge}}C_{\mathrm{in}}C_{\mathrm{out}}(m+1)^{d}$, with factor 3 stemming from 1 addition and 2 multiplications in the inner loop of each kernel and  $C_{\mathrm{in}}$ and $C_{\mathrm{out}}$ is the number of input and output channels, respectively; \textit{(iii)} for scatter operations and the bias term, the FLOPs count is $(N_{\mathrm{edge}}+N_{\mathrm{node}})C_{\mathrm{out}}$, where $N_{\mathrm{node}}$ is the number of nodes. In total, we have 
\begin{eqnarray} \label{eq:GCN_FLOPs}
\mathrm{FLOPs} & = & N_{\mathrm{edge}}(m+1)^{d}(3C_{\mathrm{in}}C_{\mathrm{out}}+7d)
\nonumber \\ & + & (N_{\mathrm{edge}}+N_{\mathrm{node}})C_{\mathrm{out}}
\end{eqnarray}
For fully connected layers, in both conventional CNNs and GCNs, we compute FLOPs as \cite{molchanov2016pruning} $\mathrm{FLOPs} = (2I-1)O$,
 where $I$ is the input dimensionality and $O$ is the output dimensionality. As to the number of parameters, for each convolution layer in both CNNs and GCNs, it is $(C_{\mathrm{in}}K^{2}+1)C_{\mathrm{out}}$, while in fully connected layers, it is $(C_{\mathrm{in}}+1)C_{\mathrm{out}}$. As shown by \eqref{eq:GCN_FLOPs}, FLOPs of graph convolution  depend on the number of edges and nodes. Since the size of input graph varies per dataset, we opt to report representative results from N-Caltech101 in Table \ref{t:flops}. G-CNNS and RG-CNNs have the smaller number of weights and require the less computation compared to deep CNNs. The main reason is that the graph representation is compact, which in turn  reduces the amount of data that needs to be processed. For N-Caltech101, the average number of nodes of each graph is  1000, while grouping events to 2-channel image makes the input size equal to 86,400.

\subsection{Action Recognition}\label{sec:recognition}
      
\begin{figure*}
\vspace{-0.15in}
  \centering
  \includegraphics[width=7.2in]{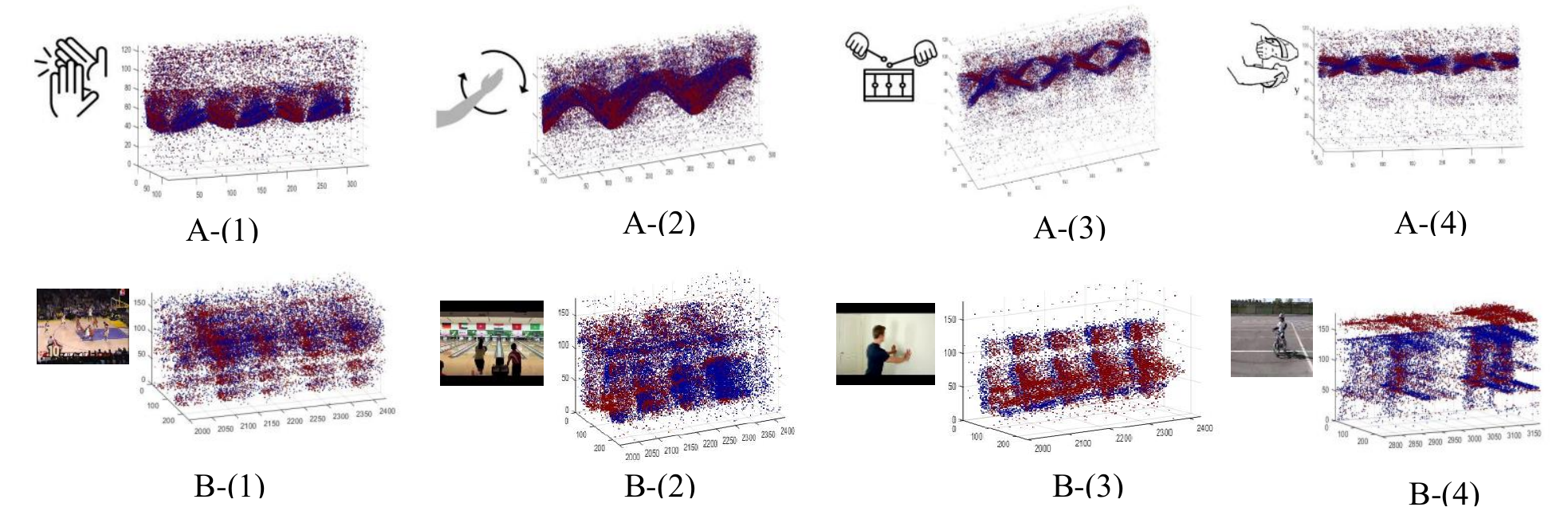}
  \vspace{-0.1in}
  \caption{Visualization of samples from DVS128 Gesture Dataest   \cite{amir2017low} and UCF101-DVS  \cite{soomro2012ucf101}. (A) DVS128 Gesture Dataset : A-1: hand clap; A-2: right hand rotation clockwise; A-3: air drums; A-4: forearm roll. (B) UCF101-DVS: B-1: basketball dunk; B-2: bowling; B-3: wall pushups; B-4: biking }
  \label{f:ibm_ucf}
\vspace{-0.15in}
\end{figure*}

\textbf{Datasets:} Previous work on neuromorphic vision sensing for action recognition evaluates on the  DVS128 Gestures Dataset \cite{amir2017low} and posture dataset \cite{zhao2014feedforward}.  DVS128 Gesture Dataset comprises 1,342 instances of 11 hand and arm gestures, while the posture dataset includes only three human actions, namely, ``bend'', ``sit/stand'' and ``walk''. Both datasets were collected from an experimental setting environment with a monotonous background, and relative to equivalent datasets for APS-based evaluation datasets, both are modest in their size and class count; as such, they cannot represent complex real-life scenarios and are not robust to evaluation for advanced algorithms. Moreover, previous work \cite{amir2017low, zhao2014feedforward, peng2016bag, wang2019space}  already achieves high accuracies on them. This is why, it is necessary to establish larger and more complex datasets for the evaluation of our proposal and for future proposals on NVS-based action recognition.

We provision two new neuromorphic event datasets, namely UCF101-DVS and HMDB51-DVS. Both datasets were  respectively captured from playbacks of the UCF101  \cite{soomro2012ucf101} and HMBD \cite{kuehne2011hmdb} datasets, which are well established datasets for the evaluation of action recognition  in the APS domain. UCF101 comprises 13,320 videos of 101 different human actions, while HMDB51 includes 6,766 videos with 51 human action categories. Of relevance is the work of Hu \textit{et al.} \cite{hu2016dvs} which previously recorded UCF50 by displaying existing benchmark videos to stationary neuromorphic vision sensors under controlled lighting conditions. We follow a  recording procedure similar to that of\cite{hu2016dvs} to wholly capture \textit{remaining} of UCF101 and HMDB51. Displayed videos are recorded by a neuromorphic vision sensor DAVIS240c that is adjusted to cover the region of interest on the monitor. Our captured datasets are the largest neuromorphic datasets for action recognition, and will be released to the public domain as a contribution of the paper once the review process is complete.

%  The details of Spatial feature learning module architecture for this module is $\mathrm{GConv(2, 32)}$ $\longrightarrow$ $\mathrm{Pool(2, 2)}$ $\longrightarrow$ $\mathrm{GConv(32, 64)}$ $\longrightarrow$ $\mathrm{Pool(4, 3)}$ $\longrightarrow$ $\mathrm{GConv(64,128)}$. And all pooling player in this module is Max-Pooling strategy. 

% As to Graph2Grid module, the parameter of this module is as following: given a graph that each node has 128 feature, we set cluster size as $8 \times 6$, so that we can obtained a $30 \times 30 \times 128$ frame for each graph given the sensor resolution is $240 \times 180$.

\textbf{Implementation Details:} We present our results on action recognition in  Table \ref{t:gesture} and Table \ref{t:ActionRes}, where the total number of graphs constructed from each event stream $S$ is set to either 8 or 16. Events within  $T_\mathrm{vol}=1/30$ seconds are constructed into one spatial graph, where individual nodes are connected to their nearest neighbor. Spatial features
are learned using our proposed residual graph CNNs (RG-CNN)  where two residual blocks are stacked, each followed by a graph max-pooling layer. Specifically, for DVS128 Gesture Dataset \cite{amir2017low} we use the architecture:  $\mathrm{Res_{g}(1,64)}$$\longrightarrow$$\mathrm{MaxP_g(2,2)}$$\longrightarrow$$\mathrm{Res_{g}(64,128)}$ $\longrightarrow$$\mathrm{MaxP_g(4,4)}$. Similarly, for  UCF101\--DVS and HMDB51\--DVS we use three residual blocks, and the architecture is: $\mathrm{Res_{g}(1,32)}$$\longrightarrow$$\mathrm{MaxP_g(2,2)}$$\longrightarrow$$\mathrm{Res_{g}(32,64)}$ $\longrightarrow$$\mathrm{MaxP_g(4,3)}$ $\longrightarrow$$\mathrm{Res_{g}(64,128)}$ $\longrightarrow$$\mathrm{MaxP_g(8,6)}$.  For the temporal feature learning module, we explore three types of architectures as described in Section \ref{sec:temporallearning}:

\textit{1) Plain 3D: }We first consider a series of consecutive 3D convolutional and pooling layers, where each intermediate convolution layer is followed by batch normalization layer and a ReLU activation function. We use $\mathrm{Conv_{3D}}(c_\mathrm{in}, c_\mathrm{out})$ to denote traditional 3D convolutional layers with batch normalization and activation functions, where $c_\mathrm{in}$ and $c_\mathrm{out}$ are the number of input and output channels respectively. 3D max pooling and global average pooling are  denoted as $\mathrm{Pool3D}$ and $\mathrm{GlobAvgP}$ respectively, fully connected layers as $\mathrm{FC}$ and task classes as $Q$. Plain 3D convolution architectures are thus  represented as follows: $\mathrm{Conv_{3D}(128,128)}$ $\longrightarrow$ $\mathrm{Pool3D}$ $\longrightarrow$ $\mathrm{Conv_{3D}(128,256)}$ $\longrightarrow$ $\mathrm{Pool3D}$ $\longrightarrow$ $\mathrm{Conv_{3D}(256,512)}$ $\longrightarrow$ $\mathrm{Pool3D}$ $\longrightarrow$ $\mathrm{Conv_{3D}(512,512)}$ $\longrightarrow$ $\mathrm{Pool3D}$ $\longrightarrow$ $\mathrm{GlobAvgP}$ $\longrightarrow$ $\mathrm{FC}(Q)$. With the notation $(h,w,t)$ denoting height, width and time dimensions, we note that the    kernel size and stride in every convolution layer is $(3,3,3)$ and $(1,1,1)$ respectively, and the window size and stride of all 3D max pooling layers is $(2,2,2)$, expect for the first pooling layer, where the stride is $(2,2,1)$ to ensure that  temporal downscaling  is not aggressive early on. 

\textit{2) Inception-3D(4):}  We next consider an Inception-3D  architecture, comprising a series of four consecutive I3D blocks. In order to ensure  that  temporal feature learning is not  bottlenecked, we  restrict the number of I3D blocks to four. Similar to \cite{carreira2017quo}, our implementation of the  I3D block is a concatenation of four streams of convolutional layers with varying kernel sizes. Where we use the shorthand   $\mathrm{Inc_b(c_\mathrm{in},c_\mathrm{out})}$ to denote each $b$-th  I3D block, we setup our architecture as:  $\mathrm{Inc_1(128,480)}$ $\longrightarrow$ $\mathrm{Pool3D}$ $\longrightarrow$ $\mathrm{Inc_2(480,512)}$ $\longrightarrow$ $\mathrm{Pool3D}$ $\longrightarrow$ $\mathrm{Inc_3(512,512)}$$\longrightarrow$ $\mathrm{Pool3D}$ $\longrightarrow$ $\mathrm{Inc_4(512, 512)}$ $\longrightarrow$ $\mathrm{Pool3D}$ $\longrightarrow$ $\mathrm{GlobAvgP}$  $\longrightarrow$ $\mathrm{FC}(Q)$.  The number of output channels of the $n$-th convolutional layer for the $s$-th stream is labelled as $c_{\mathrm{out}}[s][n]$, and the number of output channels per convolutional layer for each I3D block is: $[[128],[128,192],[32,96],64]$], $[[192],[96,208],[16,48],64]]$, $[[160],[112,224],[24,64],64]]$ and $[[128],[128,256],[24,64],64]]$.  

\textit{3) Residual 3D:} Finally, we consider 3D residual CNNs, where we effectively replace the I3D block with a 3D residual block. The 3D residual block design for temporal feature learning is illustrated at the bottom right of Fig. \ref{f:framework};   essentially, there are two 3D convolutional layers in  the base stream of the block, with a  non-linear residual connection from the input of the first to the output of the second layer. We can  define a 3D residual block as $\mathrm{Res}(c_\mathrm{in},c_\mathrm{inter},c_\mathrm{out})$, where $c_\mathrm{inter}$ represents the number of input channels to the second convolutional layer in the base stream and $c_\mathrm{in}$ and $c_\mathrm{out}$ are the respective number of input and output channels  to the residual block.   The 3D residual CNN is defined as follows: $\mathrm{Res(128,256,512)}$ $\longrightarrow$ $\mathrm{Pool3D}$ $\longrightarrow$ $\mathrm{Res(512,512,1024)}$ $\longrightarrow$ $\mathrm{Pool3D}$ $\longrightarrow$ $\mathrm{GlobAvgP}$ $\longrightarrow$ $\mathrm{FC}(Q)$. Again, denoting  $(h,w,t)$ as the height, width and time  dimensions, the kernel size is $(3,3,3)$ and stride is $(1,1,1)$ for all convolutional layers in the base stream.

In all of our tests, sampled graphs are spatially scaled by  random sampling factors within  $[0.8,1],$ and are randomly left-right flipped with a probability of $0.5$. For all of our reported results, we train using the Adam optimizer for $150$ epochs, with batch sizes respectively set to  $32$  and $16$ for $S=8$ and $S=16$.  The  learning rate is set to $0.001$, with stepwise decay by a factor of $0.1$ after $60$ epochs.

\textbf{Reference Networks:} We compare action recognition results of our proposed  RG-CNN + Plain 3D, RG-CNN + Incep. 3D(4) and RG-CNN + Res. 3D models   with previous proposals for the APS domain, where we  repurpose their use to the NVS domain by maintaining the spatial coherence of events to pass them as input frames. As external benchmarks, we include C3D \cite{tran2015learning}, I3D \cite{carreira2017quo}, 3D ResNet with 34 layers \cite{he2016deep}, P3D with 63 layers \cite{qiu2017learning}, R2+1D \cite{tran2018closer} and 3D ResNext with 50 layers \cite{hara2018can}. In contrast to our framework, these aforementioned proposals are entirely grid-based, and we construct independent frames for their use by summing events within a $1/30 $ seconds duration at each spatial position of the  NVS sensor. In this way,  resulting event frames are represented by two channels, where   ON and OFF events are grouped independently, and in order to align event maps with the number of input graphs utilized
in our framework, we produce $S=8  $ or $S=16$ sampled frames for each input volume of events. To avoid over-fitting during training, we supplement training with data augmentation, where we normalize the input and re-size the input frames such that the smaller side is $128$ (178 for P3D, 256 for I3D) and keep the aspect ratio, and use a random cropping to acquire appropriately sized inputs, and cropped volumes are randomly left-right flipped with a probability of $0.5$. We randomly initialize the parameters of all models and use stochastic gradient descent with momentum set to $0.9$, and  learning rate initialized at $0.01$ with a decay factor  of $0.1$ every $50$ epochs.

\textbf{Results:} We first evaluate our method on the DVS128 Gesture Dataset, and compare with both recent state-of-theart methods and reference networks. The results are shown in Table \ref{t:gesture}, and for all recent methods, considered event recording durations are set to 0.25 and 0.5 seconds. We follow the same set up to set the number of graphs, enabling a fair comparison. Examining the results,  we find LSTM-based methods \cite{sainath2015convolutional} to be outperformed by others, and we attribute this to the fact  that  LSTMs regard event streams as pure temporal sequences and only learn temporal features from events, without encoding spatial dependencies. In contrast, PointNet-based methods \cite{qi2017pointnet, qi2017pointnet++, wang2019space} are more accurate, and consider inputs as  point clouds to learn to summarize their geometric features. With regards to reference networks, although I3D\cite{carreira2017quo} and 3D ResNet-34\ \cite{he2016deep} perform spatio-temporal feature learning, there is no explicit modelling of event dependencies as events are directly grouped into frames. As such, our proposal outperforms all existing works and reference networks on this dataset and sets a new benchmark. We attribute this to the combination of our graph representation, spatial feature learning and temporal feature learning over multiple graphs, which results in learning a more informative spatio-temporal representation of the input.

\begin{table*}[ht]
\centering
\caption{Top-1 classification accuracies on the DVS128G gestures dataset. Performance is reported for input duration with temporal depths of 0.25 and 0.5 seconds. }\label{t:gesture}
\begin{tabular}{ccc}
\hline
Method & Duration(0.25s) & Duration(0.5s)\\
\hline
LSTM \cite{sainath2015convolutional} & 0.882 & 0.865 \\
PointNet \cite{qi2017pointnet} & 0.887 & 0.902 \\
PointNet++ \cite{qi2017pointnet++} & 0.923 & 0.941 \\
Amir CVPR2017 \cite{amir2017low} & - & 0.945 \\
Wang WACV2019 \cite{wang2019space} & 0.940 &0.953\\
\hline
ResNet\_34 \cite{he2016deep} & 0.943 &0.955\\
I3D \cite{carreira2017quo} & 0.951 &0.965\\
\hline  
RG-CNN + Plain 3D & 0.954 & 0.968 \\
RG-CNN + Incep. 3D & 0.957 & 0.968 \\
RG-CNN + Res. 3D & \textbf{0.961} & \textbf{0.972 }\\
\hline            
\end{tabular}
\end{table*}

As shown in Fig. \ref{f:ibm_ucf}, DVS128 Gesture Dataset contain salient pattern differences, while UCF101-DVS comprises more complex event volumes, and as shown in Table \ref{t:gesture}, results of the best performing models on DVS128 Gesture Dataset are close to achieving complete accuracy. Therefore, we further evaluate our
algorithms on our newly introduced datasets, UCF101-DVS
and HMDB51-DVS, which contain more classes and overall
present a more challenging task for action recognition. We note that when evaluating current NVS-based methods for
action recognition on UCF101-DVS and HMDB51-DVS, the
accuracy obtainable is only around 5\%-7\%, since these methods only perform spatial (PointNet, PointNet++) or temporal (LSTM) feature learning, and thus leaning to degenerate solutions. Therefore, we focus our comparison on reference networks for these datasets.

The Top-1 recognition accuracy of all models is reported in Table \ref{t:ActionRes} for UCF101-DVS and HMDB51-DVS, where it shows that  all variants of our architecture outperform tested benchmarks. Specifically, the highest performance obtained from reference models is from I3D, while our base  model (RG-CNN + Plain 3D) outperforms I3D by 3.3\% on UCF101-DVS and by 6.1\%  on HMDB51-DVS when $S=8$. The accuracy of our models is further increased when considering the Inception-3D and  Residual-3D variants, where our model performance increases slightly due to the higher capacity of these architectures.

\begin{table*}[ht]
\centering
\caption{Top-1 classification accuracy of UCF101-DVS and HMDB51-DVS w.r.t. various model.}\label{t:ActionRes}
\begin{tabular}{ccccc}
\hline
\multirow{2}*{Model} & \multicolumn{2}{c}{UCF101-DVS} & \multicolumn{2}{c}{HMDB51-DVS}\\
 & $S=8$ & $S=16$ & $S=8$ & $S=16$ \\
\hline
C3D \cite{tran2015learning} & 0.382 & 0.472 & 0.342 & 0.417\\
%DenseNet & & 0.575 & & \\
ResNet-34 \cite{he2016deep}& 0.513 & 0.579 & 0.350 & 0.438 \\
%ResNet50\_A & 0.404 & 0.524 & 0.164 & 0.197 \\
%ResNet50\_B & 0.502 & 0.558 &  & 0.247 \\
%P3D63\_A & 0.410 & 0.496 &  &  \\
P3D-63 \cite{qiu2017learning}& 0.484 & 0.534 & 0.343 & 0.404 \\
R2+1D-36 \cite{tran2018closer} & 0.496 & 0.628 & 0.312 & 0.419 \\
%ResNext50\_A & 0.504 & 0.589 & 0.306 & 0.393 \\
ResNext-50 \cite{hara2018can} & 0.515 & 0.602 & 0.317 & 0.394 \\
I3D \cite{carreira2017quo}& 0.596 & 0.635 & 0.386 & 0.466 \\
\hline  
RG-CNN + Plain 3D & 0.629 & 0.663 & 0.447 & 0.494\\
RG-CNN + Incep. 3D & \textbf{0.632} & \textbf{0.678} & 0.452 & \textbf{0.515}\\
RG-CNN + Res. 3D & 0.627 & 0.673 & \textbf{0.455} & 0.497\\
\hline            
\end{tabular}
\end{table*}

\textbf{Complexity Analysis: }We compare the complexity of tested models, and do so with respect to the number of floating-point operations (FLOPs) and required parameter counts. For graph-based convolutional and fully-connected
layers, FLOPs and parameter counts are calculated as detailed in Section \ref{sec:recognition}. For conventional 3D convolutional layers, we compute FLOPs as $2HWT(C_{\mathrm{in}}K^{3}+1)C_{\mathrm{out}}$ multi-add operations, where $H$, $W$, and $T$  are the height, width, and temporal length of input maps,   $C_{\mathrm{in}}$ is the number of input feature channels, $K$ is the kernel size, and $C_{\mathrm{out}}$ is the number of output channels. Using similar notation, parameter counts of conventional 3D convolutional layers are calculated as $(C_{\mathrm{in}}K^{3}+1)C_{\mathrm{out}}$.    FLOPs of graph convolutions depend on edge and node counts (see Section \ref{sec:recognition}), and we specifically report results for UCF101-DVS in Table \ref{t:complexity}. For each sample, $16$ graphs are sampled as inputs to the spatial feature learning module, and FLOPs in respective modules are the averages over the whole of UCF101-DVS.
Our results show how graph convolutions can manage with smaller or comparably sized input volumes relative to all reference models. As for complexity, though our models require more floating-point operations when compared to P3D\--63 and ResNext\--50, they achieve better performance on all  three datasets. On the other hand, accuracies of  I3D are close to ours while requiring complexities which are  two to three times higher.

\begin{table}[ht]
\centering
\caption{Comparison of models w.r.t. complexity (GFLOPs) and size of architecture parameters.}\label{t:complexity}
\begin{tabular}{ccc}
\hline
Model & GFLOPs & Parameters($\times10^{6}$)\\
\hline
C3D \cite{tran2015learning}& 39.69 & 78.41\\             %C3D & 38.5 & 72.9\\
ResNet-34 \cite{he2016deep}& 11.64 & 63.70\\        %ResNet\_34 & 36.7 & 63.5\\
P3D-63 \cite{qiu2017learning}& 8.30 & 25.74\\
R2+1D-36 \cite{tran2018closer} & 41.77 & 33.22\\          %R2+1D  & 152.4 & 63.6\\
ResNext-50 \cite{hara2018can}& 6.46 & 26.05\\
I3D \cite{carreira2017quo}& 30.11 & 12.37\\            %I3D & 107.9 & 12.1\\
\hline  
RG-CNN + Plain 3D & 12.46& 6.95\\
RG-CNN + Incep. 3D & 12.39 & 3.86\\
RG-CNN + Res. 3D & 13.72 & 12.43\\
\hline            
\end{tabular}
\end{table}   

\subsection{Action Similarity Labeling}\label{sec:similarity}
Action similarity labeling is a binary classification task wherein  alignments of action pairs are predicted. In other words, models are required to learn to evaluate the similarity of actions rather than recognize particular actions. The challenge of action similarity labeling lies in that the actions of test sets  belong to separate classes and are not available during training\cite{kliper2011action}. That is to say, training does not provide an opportunity to learn  actions presented at test time. To the best of our knowledge, as of yet there is no work on similarity detection in the neuromorphic domain, and no existing dataset can be used for evaluation. We use the ASLAN \cite{kliper2011action} dataset which comprises 3,631 samples from 432 different action classes. Using a similar  setting to the one described in Section IV-B, we captured an equivalent neuromorphic dataset  ASLAN-DVS  to be publicly provisioned for relevant research.

\textbf{Training Details:} We use the ``View-2'' method as detailed in  \cite{kliper2011action} to split samples into 10 mutually exclusive subsets, where each subset contains 600 video pairs, with 300 to be classified as "similar" and 300 to be classified as "not similar". We report our results  by averaging scores of 10 separate experiments in a leave-one-out cross validation scheme. In this application, we used models trained for action recognition as feature extractors, and  extracted  $L_{2}$-normalised output features from the  last $\mathrm{ GlobalAvgP}$ and  $\mathrm{Pool3D}$ layers to acquire two distinct types of representation. Similar to\cite{kliper2011action}, we independently compute 12 different distances for said features and for every pair of actions.  Finally, a support vector machine with a radial basis kernel is trained to classify whether action pairs are of similar or different activities. As baselines, we consider the performance of  reference architectures detailed in Sec. \ref{sec:recognition}, where features are  extracted as the outputs of the last two layers, and  classifications are performed by support vector machines. The complexity of our proposed spatio-temporal  feature learning and other reference models remain the same as in Section \ref{sec:recognition}.

In Table \ref{t:SimilarityRes} we report the performance of different models as
measured accuracies and areas under ROC curves (AUC). Our RG-CNN + Incep. 3D
framework outperforms state-of-the-art results acquired from I3D by 2.6\%
on accuracy and 3.1\% on AUC, %We also plot ROC curve in Fig.\ref{f:auc}
which clearly indicates that  graph-based models are better suited for feature learning for the purposes of action similarity labeling.

%\begin{figure}
%\centering
%\vspace{-0.1in}
%\includegraphics[width=2.8in]{figure/AUC.pdf}
%\vspace{-0.1in}
%\caption{Action similarity labeling result. ROC curves of proposed and reference networks evaluated on ASLAN-DVS.}
%\vspace{-0.1in}
%\label{f:auc}
%\end{figure}

\begin{table}[ht]
\centering
\caption{Action similarity detection performance on ASLAN-DVS \cite{kliper2011action}  w.r.t. tested models.}\label{t:SimilarityRes}
\begin{tabular}{ccc}
\hline
Model & Acc. & AUC\\
\hline
ResNet-34 \cite{he2016deep}& 0.605 & 0.643\\
P3D-63 \cite{qiu2017learning}& 0.598 & 0.638\\
R2+1D-36 \cite{tran2018closer} & 0.615 & 0.652\\
ResNext-50 \cite{hara2018can}&  0.605 & 0.643\\
I3D \cite{carreira2017quo}& 0.623 & 0.659\\
\hline  
RG-CNN + Plain 3D & 0.635 &0.674\\
RG-CNN + Incep. 3D(4) &\textbf{ 0.649} & \textbf{0.690}\\
RG-CNN + Res. 3D & 0.641 & 0.684\\
\hline            
\end{tabular}
\end{table}

\section{Conclusion}\label{sec:conclusion}
In this work we develop an end-to-end trainable graph-based feature learning framework for neuromorphic vision sensing. We first represent neuromorphic events as graphs, which are explicitly aligned with the compact and non-uniform sampling of NVS hardware. We couple this with an efficient end-to-end learning framework, comprising graph convolutional networks for spatial feature learning directly from graph inputs.  We extend our framework with our Graph2Grid module that converts the graphs to grid representations for coarse temporal feature learning with conventional 3D\ CNNs.  We demonstrate how this framework can be employed for object classification, action recognition and action similarity labeling, and evaluate our framework on all tasks with standard datasets.  We  additionally propose and make available three large-scale neuromorphic datasets in order to motivate further progress in the field. Finally, our results on all datasets show  that we outperform all recent NVS-based proposals while maintaining lower complexity.

{\small
\bibliographystyle{ieee}
\bibliography{ref}
}

\

\end{document}